\documentclass[conference]{IEEEtran}
\IEEEoverridecommandlockouts
\usepackage{longtable}
\usepackage{cite}
\usepackage{amsmath,amssymb,amsfonts}
\usepackage{algorithmic}
\usepackage{graphicx}
\usepackage{textcomp}
\usepackage{xcolor}
\usepackage{booktabs}
\usepackage{multirow}
\usepackage{siunitx}
\usepackage{hyperref}
\def\BibTeX{{\rm B\kern-.05em{\sc i\kern-.025em b}\kern-.08em
    T\kern-.1667em\lower.7ex\hbox{E}\kern-.125emX}}

\usepackage{fancyhdr}
\begin{document}

\title{Contextual Emotion Estimation from \\Image Captions
{\footnotesize \textsuperscript{}} % so delete this line
\thanks{This work was supported by NSERC Discovery Grant 06908-2019.}
}

\author{
% \IEEEauthorblockN{Anonymous}
% \IEEEauthorblockA{\textit{Department} \\
% \textit{Organization}\\
% Email Address
% }
% \and
\IEEEauthorblockN{Vera Yang, Archita Srivastava, Yasaman Etesam, Chuxuan Zhang, Angelica Lim}
\IEEEauthorblockA{\textit{School of Computing Science} \\
\textit{Simon Fraser University}\\
Burnaby, Canada \\
\{veray, archita\_srivastava, yetesam, chuxuan\_zhang, angelica\}@sfu.ca}
% \and
% \IEEEauthorblockN{Anonymous}
% \and
% \and
% \and
}

\maketitle
\thispagestyle{fancy}

\begin{abstract}
% Traditional methods to estimate people’s emotions in images have focused primarily on the face and body pose. However, previous research has proven that relying only on this information is inadequate. On the other hand, Large Language Models (LLMs) possess information about various aspects of humans. But how well do LLMs perceive human emotions? And which parts of the information enable them to determine emotions? 

Emotion estimation in images is a challenging task, typically using computer vision methods to directly estimate people's emotions using face, body pose and contextual cues. In this paper, we explore whether Large Language Models (LLMs) can support the contextual emotion estimation task, by first captioning images, then using an LLM for inference. First, we must understand: how well do LLMs perceive human emotions? And which parts of the information enable them to determine emotions? One initial challenge is to construct a caption that describes a person within a scene with information relevant for emotion perception. Towards this goal, we propose a set of natural language descriptors for faces, bodies, interactions, and environments. We use them to manually generate captions and emotion annotations for a subset of 331 images from the EMOTIC dataset. These captions offer an interpretable representation for emotion estimation, towards understanding how elements of a scene affect emotion perception in LLMs and beyond. Secondly, we test the capability of a large language model to infer an emotion from the resulting image captions. We find that GPT-3.5, specifically the text-davinci-003 model, provides surprisingly reasonable emotion predictions consistent with human annotations, but accuracy can depend on the emotion concept. Overall, the results suggest promise in the image captioning and LLM approach.

% Large language models have been a hot topic in recent years. The recent improvements in these models have increased the accuracy of many tasks, such as question answering or even image captioning. \yasaman{On the other hand emotion estimation is an important task..} that In this paper, we examine how these models can perceive human emotions and understand the impact of context and surroundings on human emotions. \yasaman{mention chatgpt?}

\end{abstract}

\begin{IEEEkeywords}
Large language model, emotion estimation, image captioning, context, ChatGPT, GPT-3.5
\end{IEEEkeywords}

\begin{figure*}[t]
\centerline{\includegraphics[width = 18cm]{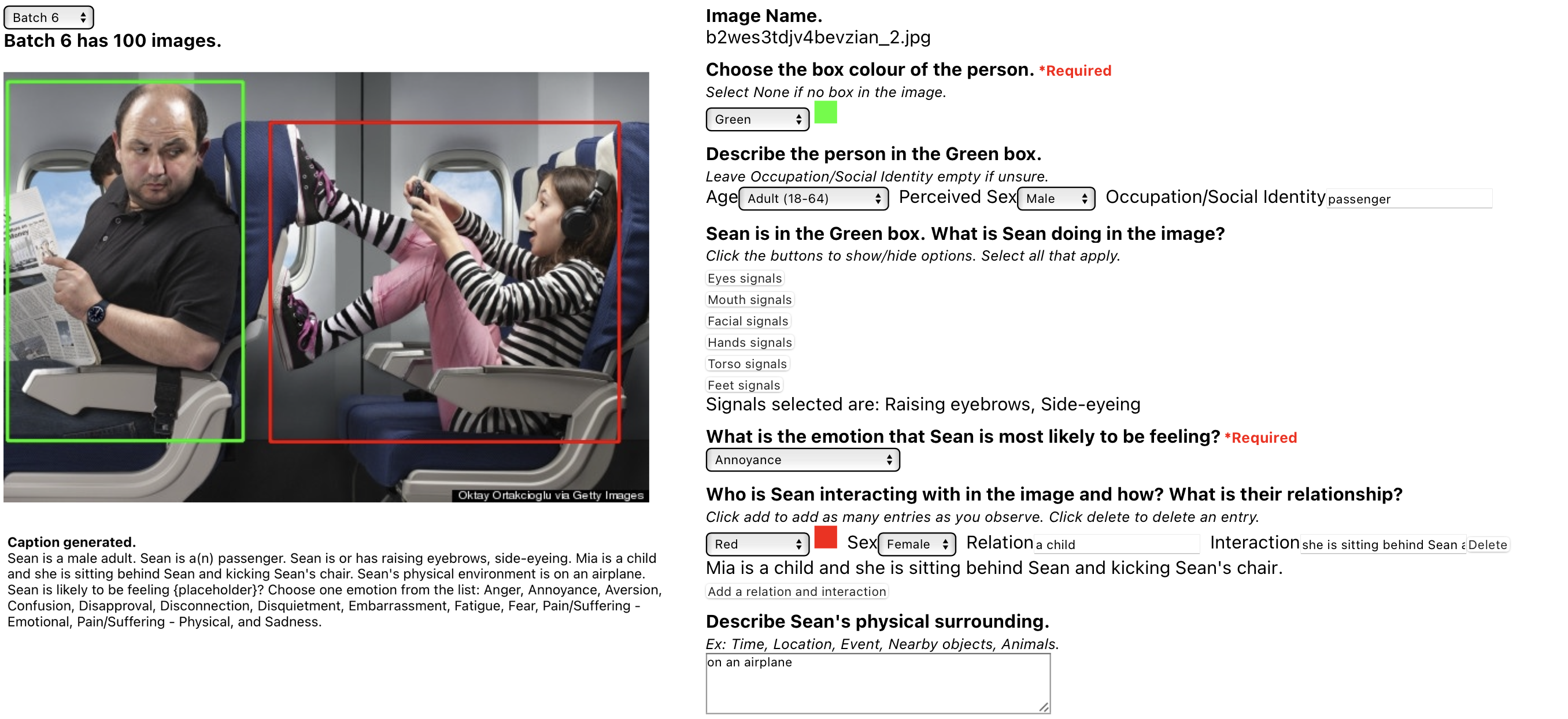}}
\caption{Manual Annotation for the given image produces the following caption: 
\textit {Sean is a male adult. Sean is a(n) passenger. Sean is or has raising eyebrows, side-eyeing. Mia is a child and she is sitting behind Sean and kicking Sean's chair. Sean's physical environment is on an airplane.}}
\label{WebAnnotation}
\end{figure*}

\section{Introduction}
\emph{``She sat in a hospital hallway, with an empty stare and slumped shoulders."} How does this person feel? Writers have long known that describing a scene with carefully selected words, without specifically naming the emotion, is an effective way of moving their reader. The ability to place ourselves in the shoes of another underlies our ability to infer their emotion, towards taking socially appropriate and empathetic actions. Similarly, a photo can capture the emotion of a person in a scene. Automatic emotion estimation systems based on images or videos have the potential to facilitate better human-machine interaction, yet performance in the wild is still poor~\cite{barrett2019emotional}. 

Many emotion recognition studies focus on using facial \cite{pantic2000expert} or body\cite{schindler2008recognizing} features. The context in which emotions are expressed can also affect the perception of emotions \cite{barrett2011context, barrett2017emotions, calbi2017context,7284842}, whether the face is visible or covered in the image. As a result, the context-based emotion recognition task was introduced. It was elaborated with the introduction of the EMOTIC dataset \cite{kosti2019context}, and there has been an increased focus on improving accuracy on this task \cite{le2022global, mittal2020emoticon, wang2022context}. These models utilize a variety of inputs beyond facial data by including, for examples, body posture and context, which encompass factors such as the presence of other humans or environmental aspects. Context-based emotion recognition in audio-visual media has also been studied \cite{dudzik2020exploring}, but how exactly specific cues contribute to the detected emotion remains a relatively unexplored area \cite{shin2022contextual}.

In recent years, large language models (LLMs) have emerged as a hot topic in the field of Natural Language Processing (NLP). This growth can be attributed to the introduction of transformers in 2017 by Vaswani et al. \cite{vaswani2017attention}, which provided a more efficient way of processing sequences of data. Subsequently, other researchers have introduced various methods based on transformer encoder/decoder structures and different pre-training techniques \cite{radford2019language, devlin2018bert, brown2020language}. These approaches have allowed sophisticated language models to perform a range of tasks with high accuracy and efficiency. 
These improvements in LLMs paved the way not only to the improvements in NLP problems, but also to many multi-modal problems such as Visual Question Answering \cite{antol2015vqa}, and Caption Generation \cite{vinyals2015show}. 
The ability of these models to understand human language and store data in their extensive neural network attributed this success. At the same time, to what extent these language models have the ability to perceive human emotion remains an open question.

%We combined Pain and Suffering into one label, and also separated them into Emotional Pain/Suffering versus Physical Pain/Suffering. As a result, our labels are: Anger, Annoyance, Aversion, Confusion, Disapproval, Disconnection, Disquietment, Embarrassment, Fatigue, Fear, Pain/Suffering - Emotional, Pain/Suffering - Physical, and Sadness. 
In this study, we aim to answer the following questions: how well do LLMs perceive human emotions? And which parts of the information enable them to determine emotions? We first created an annotation interface that allows for annotating images with various factors related to emotion, such as physical signs, social interactions, environmental cues, and demographic information. Using this information, we created an image caption describing a person's facial expressions and body poses, their social contexts with other people in an image, and their environmental surrounding. We then passed the image caption to a GPT-3.5\footnote{https://platform.openai.com/docs/models/gpt-3-5} model to predict an emotion from the text description only.

% We also use clip\cite{radford2019language} to create these descriptions and pass it to the GPT3 model and find the predicted emotion.
We conducted an emotion prediction experiment using full image captions, followed by two ablation studies using cropped image captions. For the ablation studies, we altered our image captions by selectively removing certain types of contextual information, such as social interactions and environmental features. Through the ablation studies, we aim to evaluate the impact of each type of data on the emotion detection output. To summarize, our contributions are:
\begin{itemize}
    \item Compiling a set of physical signals for each of our emotion labels using LLMs and a writer's thesaurus \cite{puglisi2019emotion}.
    \item Developing an interface to annotate the image data and collected emotion labels and description showing the physical signals, human interactions, and environmental features for each person.
    \item Providing an initial analysis on GPT-3.5's ability to predict human emotion from image captions, and how well it can predict the emotion given information on physical signals and contextual information.
    % \item Comparing the output of GPT-3 between human generated descriptions and descriptions generated using CLIP
    \item Analyzing the importance of context on how large language models perceive emotions, and how different types of context will affect the prediction.
\end{itemize}

% \section{Related Work}
% \subsection{context-based emotion recognition}
% \subsection{large language models}
% \subsection{caption generation - clip}
% \subsection{different emotion category labels}
\section{Methodology}

\begin{table*}[t]
\caption{Physical Signals Used in Annotated Image Captions}
\begin{center}
\begin{tabular}{p{3cm}|p{14cm}}
\hline
\textbf{Signal Categories}&\textbf{Physical Signals} \\
\hline
Eyes Signals & Closed eyes, Frowning, Staring off into the distance, Furrowed eyebrows, Glaring, Side-eyeing, Averted gaze, Looking up, Squeezing eyes shut, Rolling eyes, Avoiding eye contact, Looking away, Lowered eyebrows, Empty stare, Raising eyebrows, Squinting eyes, Looking sideway, Eyebrows squishing together, Eyes wide open, Looking down, Peeking, Eyes are damp and bright, Staring down at the ground, Unfocused gaze, Downcast eyes, Raising one's eyebrows, Gaze clouding, Glassy stare, Refusing to look, Glancing as if looking for answers, Eyes are cold, Side look \\
 \hline
 Mouth Signals & Gritting teeth, Open mouth, Mouth wide open, Clenched jaw, Downturned mouth, Biting lips, Lips that flatten, Smiling, Sticking tongue out, Curling lip, Poking one's tongue into the cheek, Yawning, Smirking, Breathing excessively, Biting finger, Pressing lips tight, Grimacing \\
\hline
Facial Signals & High chin, Flat expression, Crying, Chin dipping down, Resting one side of face on hand, Resting chin on hand, Tilting one's head to the side, Smelling oneself, Hiding face in arms, Tilting head downward, Leaning head on hand, Using something to hide face, Wrinkling nose, Resting forehead on hand, Puffing out the cheeks, Tilting head upward, Cheek resting on own palm, Expression that appears pained\\
\hline
Body Signals & Sitting, Bending down, Bent spine, Bending forward, Visible sweating, Leaning back, Hunched shoulders, Leaning forward, Body posture that loosens or collapses, Lying flat, Bracing against a wall, Chest caving, Bending forward and laying head on arms, Shoulders slumping or curling forward, Leaning on an object, Falling onto ground, Tight shoulders, Naked body, Tense posture, Picking fights, Body freezing in place, Turning body away, Curling up body into a ball, Stiff posture, Slouching on an object, Hood over the head, Throwing up, Covering oneself with something, Keeping one's back to a wall, Sliding down in a chair, Slouching and leaning on objects, Visible tension in the neck, Hands covering face, Pointing fingers, Pulling a hood over the head, Rubbing the affected area of one's body \\
\hline
Hands Signals & Hand up in air, Folding arms across the chest, Hand wiping tears, Covering face with hands, Thumbs down, Pressing a fist to the mouth, Throwing things, Pointing middle finger, Palms open, Curling fingers, Rubbing one's forehead or eyebrows, Hands in pockets, Crossed arms, Squeezing nose, Grabbing onto someone, Hand on own chest, Hand resting on forehead, Hands on both sides of head, Rubbing the eyes, Grabbing own hair, Clenched fists, Pointing finger, Scratching at cheek or temple, Rubbing the foot, Arms reaching out, Gripping something and knuckles going white, Palms covering forehead, Palms up facing outward, Hands on the hip, Rubbing temples, Rubbing the back of the neck, Hands covering ears, Rubbing the back, Taking off eyeglasses, Hand on neck, Throwing hands up in the air, Wiping tears, Rubbing the shoulder, Sweeping hand across the forehead to get rid of sweat, Clapping palms together, Hand covering mouth, Rubbing the chest, Hands curling around their body, Rubbing the nose, Clutching the stomach, Hand touching the lips, Nervous hand gestures \\
\hline
Feet Signals & Knees pulling together, Bringing the feet together, Squatting \\
\hline

\hline
% \multicolumn{3}{l}{$^{\mathrm{a}}$Sample of a Table footnote.}
\end{tabular}
\label{signal_table}
\end{center}
\end{table*}

The ultimate goal of our research is to explore whether automatic emotion estimation of people in images could be implemented by first captioning images appropriately, then feeding the caption to a large language model for inference on the text. Our approach is comprised of three steps: a) generating a large list of physical signals used for writing about emotion, b) annotating images using these signals along with questions about demographic information, interaction, and environment, and c) using a large language model to predict an emotion based on the image caption.

\subsection{Generation of Physical Signals} Because existing algorithms perform relatively poorly on negative labels compared to positive ones (e.g. 14.5\% vs. 40.3\% \cite{kosti2019context}), we focused on the 13 negative emotion labels from the EMOTIC dataset \cite{kosti2019context}: Anger, Annoyance, Aversion, Confusion, Disapproval, Disconnection, Disquietment, Embarrassment, Fatigue, Fear, Pain, Sadness, and Suffering. To mitigate an annotator's potential confusion with the labels Pain and Suffering, we merged and replaced these labels with Pain/Suffering - Emotional and Pain/Suffering - Physical. 

As a first step, we generated descriptions of physical signals indicative of our 13 target emotions. We used an Emotion Thesaurus, ``A Writer’s Guide to Character Expression” by Becca Puglisi and Angela Ackerman \cite{puglisi2019emotion}, which provided a range of physical signals associated with commonly recognized emotions including Anger, Annoyance, Confusion, Embarrassment, Fear, and Sadness. For emotions not listed in the book, we utilized the Large Language Models ChatGPT and GPT-3.5 to generate a list of physical descriptions or expressions associated with each emotion label. The prompts used to generate the physical descriptions were of the form, \textit{``List physical cues/physical expressions that would indicate the emotion of `disapproval' in an image."} and \textit{``Give a list of facial expressions/physical descriptions/physical movements that might indicate that a person is feeling `fatigued'."} 

The generated descriptions were then filtered and combined to create a comprehensive set of physical signals for our set of emotion labels. This resulted in a total of 222 distinct physical signals that could indicate the emotion an individual is experiencing in an image. It should be noted that the remainder of the study did not assume that any specific physical signals were associated with any particular emotion. 

\subsection{Annotations}
The interface shown in Fig. \ref{WebAnnotation} was created to facilitate image annotation.
To assess a large language model's ability to predict human emotions from images, we annotated a set of images from the EMOTIC dataset\cite{kosti2019context}. These images contained bounding boxes of different colours surrounding the people in the scene. This allowed us to focus on one person (e.g. marked with a ``red" bounding box) at a time within an image. 

During the annotation process, both physical signals and contextual components were considered. To make the annotation process easier, we divided the physical signals into multiple categories based on body parts, and annotators could use checkboxes to select relevant descriptions. The annotator could also tag the person within a bounding box with various attributes, including their perceived age group, perceived sex, and social identity or occupation. 

Finally, contextual information including factors such as their social interactions (e.g. alone, surrounded by people), social relationships with others in the image (e.g. mother and daughter, husband and wife), and their environmental setting could be input into an open text box. In the end, the annotation interface generated an appropriate image caption based on all the chosen tags (e.g. creating a sentence with a first name), allowing the human annotator to double-check the caption before saving their work. 

Out of 222 physical signals proposed to the annotators, 153 were ultimately used to describe the images in the dataset in this study and are reported in Table I. A full listing of interactions and environmental contexts derived from annotators is provided in Supplementary Materials.

\subsection{Assessing Prediction Abilities of Large Language Models}
Once the annotations were complete, GPT-3.5 was used to predict emotion labels with the help of a prompt. The prompt was structured to elicit single emotion prediction when presented with an image annotation. 

The prompt was as follows (considering the annotation from Fig. \ref{WebAnnotation}): \textit{"Sean is a male adult. Sean is a(n) passenger. Sean is or has raising eyebrows, side-eyeing. Mia is a child and she is sitting behind Sean and kicking Sean's chair. Sean's physical environment is on an airplane. Sean is likely feeling a high level of \{placeholder\}? Choose one emotion from the list: Anger, Annoyance, Aversion, Confusion, Disapproval, Disconnection, Disquietment, Embarrassment, Fatigue, Fear, Pain/Suffering (emotional), Pain/Suffering (physical), and Sadness."}

To evaluate the performance of the models, we compared the LLM's predictions to the ground truth of the images established by the annotators. For instance, from the manual annotation as shown in Fig.~\ref{WebAnnotation}, the ground truth for the person within the green bounding box was determined to be ``Annoyance". It should be noted that the ground truth labels in our study were different from that of the EMOTIC dataset \cite{kosti2019context}, which was a multi-label dataset. For this reason, it is not straightforward to evaluate the existing multilabel baseline algorithms on our dataset.

\section{Experiments}  \label{Experiments}
We conducted three experiments with our manually annotated image captions to test a large language model's ability to estimate human emotions. The first experiment used the full image captions that included all the visually contextual information in an image. We then performed two ablation studies to test the contribution of social interactions and environmental contexts in predicting emotions. Table~\ref{caption_table} shows how every image caption could differ between experiments.

\begin{table}[t]
\caption{Different Versions of an Image Caption Used in  Experiments}
\begin{center}
% \begin{tabular}{|L{0.2\linewidth} | L{0.7\linewidth}|}
\begin{tabular}{|p{0.8in}|p{6cm}|}
\hline
      \multicolumn{1}{|c|}{\textbf{Type of Caption}} &
      \multicolumn{1}{c|}{\textbf{Image Caption}} \\
\hline
Full Caption$^{\mathrm{a}}$ & Sean is a male adult. Sean is a(n) passenger. Sean is or has raising eyebrows, side-eyeing. Mia is a child and she is sitting behind Sean and kicking Sean’s chair. Sean’s physical environment is on an airplane. \\
\hline
Minus Interactions$^{\mathrm{b}}$  & Sean is a male adult. Sean is a(n) passenger. Sean is or has raising eyebrows, side-eyeing. Sean’s physical environment is on an airplane. \\
\hline
Minus Environments$^{\mathrm{c}}$ & Sean is a male adult. Sean is a(n) passenger. Sean is or has raising eyebrows, side-eyeing. Mia is a child and she is sitting behind Sean and kicking Sean’s chair. \\
\hline
\multicolumn{2}{l}{$^{\mathrm{a}}$Full captions are used in Experiment A.} \\
\multicolumn{2}{l}{$^{\mathrm{b}}$Captions without interactions are used in Experiment B.} \\
\multicolumn{2}{l}{$^{\mathrm{c}}$Captions without environments are used in Experiment C.}
\end{tabular}
\label{caption_table}
\end{center}
\vspace{-6mm}
\end{table}

% to test the significance and contribution of different types of context in predicting human emotions. Experiment One's captions included a person's facial expressions and body poses. Experiment Two's captions included a person's environmental contexts and physical surroundings in addition to face and body signals. Experiment Three's captions contained all the above information and a person's interactions with other people in an image. All the captions included a person's age and sex, as well as their social identity if it was apparent in the image. Social identity can range from a person's occupation, cultural background, to religious background. It is an important factor in identifying an emotional state because a president who is crying can be feeling much differently from a child who is crying.

\subsection{Dataset and Annotation}
% We only chose 
The image samples used in this study are from the EMOTIC dataset\cite{kosti2019context}. An image could contain one or more bounding boxes and each bounding box enclosed a person. Every person could depict multiple emotions, but only one emotion label mutually agreed upon by two annotators was picked as the ground truth and counted towards a sample for that emotion. If an image contained multiple people where one person showed \textit{Anger} while the other person showed \textit{Fear}, then the image counted towards a sample for both \textit{Anger} and \textit{Fear}. If two people in an image were showing the same emotion such as \textit{Sadness}, then the image was counted twice as a sample for \textit{Sadness}. 
%To ensure randomness, we picked the samples randomly from all images. 
Table~\ref{sample_table} shows the sample distribution. To summarize, our sample dataset\footnote{\url{https://rosielab.github.io/emotion-captions/}}consisted of:
\begin{itemize}
\item 331 unique images
\item 360 samples 
\item 360 captions generated through manual annotation
\item Two types of images: \textit{One person} and \textit{Multiple people}
\end{itemize}
All emotion categories had a sample size of 30, except for \textit{Confusion} (16) and \textit{Embarrassment} (14), and ground truth was observed by two annotators upon mutual agreement.
\begin{table}[t]
\caption{Number of Samples per Emotion Across Two Types of Images}
\begin{center}
\begin{tabular}{|c|c|c|c|}
\hline
\textbf{Emotion}&\textbf{One person}&\textbf{Multiple people}&\textbf{Total} \\
\hline
Anger & 14 & 16 & 30 \\
\hline
Annoyance & 16 & 14 & 30  \\
\hline
Aversion & 16 & 14 & 30  \\
\hline
Confusion & 12 & 4 & 16 \\
\hline
Disapproval & 18 & 12 & 30  \\
\hline
Disconnection & 18 & 12 & 30  \\
\hline
Disquietment & 15 & 15 & 30 \\
\hline
Embarrassment & 0 & 14 & 14 \\
\hline
Fatigue & 23 & 7 & 30 \\
\hline
Fear & 15 & 15 & 30  \\
\hline
Pain/Suffering - Emotional & 15 & 15 & 30 \\
\hline
Pain/Suffering - Physical & 15 & 15 & 30  \\
\hline
Sadness & 15 & 15 & 30  \\
\hline
% \multicolumn{3}{l}{$^{\mathrm{a}}$Sample of a Table footnote.}
\end{tabular}
\label{sample_table}
\end{center}
\end{table}

\begin{figure}[t]
\centerline{\includegraphics[width=9cm]{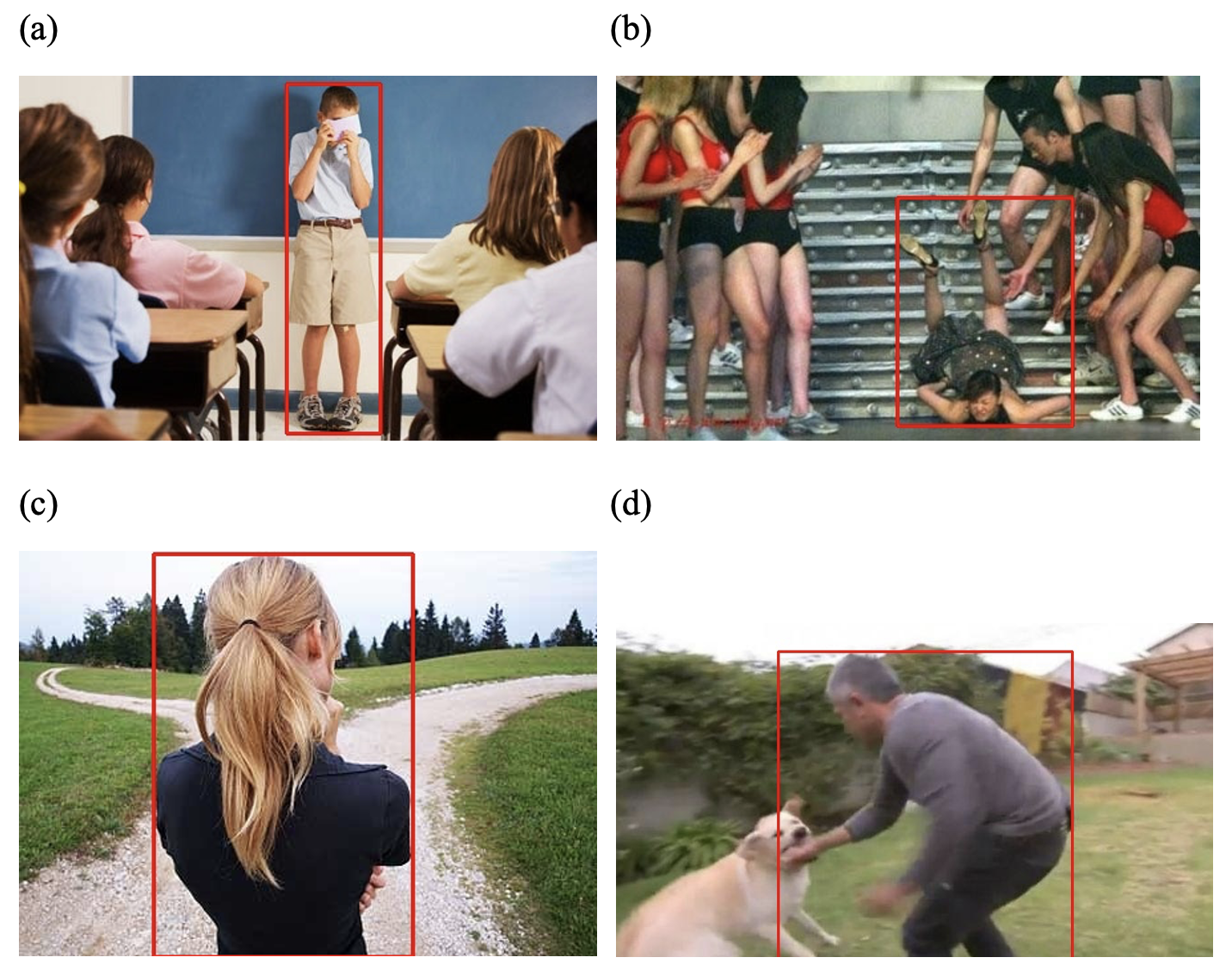}}
\caption{Images (a) and (b) are examples of how \text{Embarrassment} images required multiple people and interactions to be recognizable. Images (c) and (d) have limited facial or body signals of a person, hence we have to rely on environmental contexts to estimate an emotion.}
\label{embarrassment}
\vspace{-4mm}
\end{figure}

\begin{table*}[t]
\caption{Results of GPT-3.5 Emotion Estimation Using Captions}
\begin{center}
  \begin{tabular}{|c|ccc|ccc|ccc|}
    % \toprule
    \hline
    \multirow{2}{*}{Emotions} &
      \multicolumn{3}{c|}{Full Caption} &
      \multicolumn{3}{c|}{Minus Interactions} &
      \multicolumn{3}{c|}{Minus Environments} \\
      & {Precision} & {Recall} & {F1 Score} & {Precision} & {Recall} & {F1 Score} & {Precision} & {Recall} & {F1 Score} \\
      % \midrule
      \hline
    Anger & 0.59 & 0.77 & 0.67 & 0.55 & 0.73 & 0.63 & 0.47 & 0.70 & 0.56 \\
    \hline
    Annoyance & 0.64 & 0.23 & 0.34 & 0.67 & 0.20 & 0.31 & 0.43 & 0.10 & 0.16 \\
    \hline
    Aversion & 0 & 0 & 0 & 0 & 0 & 0 & 0 & 0 & 0 \\
    \hline
    Confusion & 0.75 & 0.19 & 0.30  & 0.75 & 0.19 & 0.30 & 1.00 & 0.13 & 0.22\\
    \hline
    Disapproval & 0.25 & 0.47 & 0.32 & 0.23 & 0.47 & 0.31 & 0.24 & 0.50 & 0.32\\
    \hline
    Disconnection & 0.25 & 0.07 & 0.11 & 0 & 0 & 0 & 0.25 & 0.10 & 0.14 \\
    \hline
    Disquietment & 0 & 0 & 0  & 0 & 0 & 0 & 0 & 0 & 0 \\
    \hline
    Embarrassment & 0.28 & 0.79 & 0.41 & 0.24 & 0.50 & 0.33 & 0.26 & 0.71 & 0.38 \\
    \hline
    Fatigue & 0.68 & 0.43 & 0.53 & 0.70 & 0.47 & 0.56 & 0.73 & 0.37 & 0.49 \\
    \hline
    Fear & 0.39 & 0.77 & 0.52 & 0.37 & 0.70 & 0.48 & 0.26 & 0.50 & 0.34 \\
    \hline
    Pain/Suffering - Emotional & 0.25 & 0.03 & 0.06 & 0 & 0 & 0 & 0.13 & 0.03 & 0.05 \\
    \hline
    Pain/Suffering - Physical & 0.86 & 0.63 & 0.73 & 0.75 & 0.40 & 0.52 & 1.00 & 0.30 & 0.46\\
    \hline
    Sadness & 0.27 & 0.87 & 0.42 & 0.21 & 0.80 & 0.33 & 0.24 & 0.80 & 0.37\\
    \hline
    \hline
    Total Accuracy & 
    \multicolumn{3}{c|}{0.39} &
      \multicolumn{3}{c|}{0.34} &
      \multicolumn{3}{c|}{0.32} \\
    % \bottomrule
    \hline
    \multicolumn{10}{l}{Random chance is 0.07.}
  \end{tabular}
  \label{stats_table}
  \end{center}
  \vspace{-6mm}
\end{table*}

\subsection{Model Parameters and Stability}
We used OpenAI's Completions API \footnote{https://platform.openai.com/docs/api-reference/completions} to provide our image captions as prompts to the GPT-3.5 model, and it returned predicted emotions through completions. The model version was \textit{text-davinci-003,} which is part of the GPT3.5 family. To ensure that the results generated from GPT-3.5 were stable and reproducible, we set the model's \textit{temperature} parameter to 0, allowing the model to give a nearly deterministic answer to every prompt. To further the reproducibility of our results, we ran the GPT-3.5 model over each caption ten times to generate a list of ten predicted emotions. We limited the emotions that GPT-3.5 could output to the 13 negative emotions that we focused on in this study. For each caption, the emotion with the maximum number of occurrences was selected as the final prediction. This prediction generation and selection process was done for all three experiments.

\subsection{Experiment A: Predicting with Full Image Captions}
Experiment A studies how accurately a large language model can predict a person's emotional state in an image given all the contextual information. A full image caption includes a person's perceived age, perceived sex, social identity if apparent, facial expression and body poses if applicable. It also includes their interactions with other people and their environmental surrounding if applicable. Table~\ref{caption_table} shows a full caption for Fig.~\ref{WebAnnotation}.

\subsection{Experiment B: Ablation Study on Interactions with People}

Experiment B studies how describing a person's interactions and relationships with other people in an image contributes to determining a person's emotional state. It used the same dataset as Experiment A. We removed all the information about a person's social interactions and relationships in an image from the full captions. Thus, the captions used in this experiment contained only perceived age, perceived sex, applicable social identity, face and body signals, and environment. Table~\ref{caption_table} shows a caption without social interaction for Fig.~\ref{WebAnnotation}. Fig.~\ref{embarrassment} (a) and (b) show two examples where social interactions may need to be considered to fully understand \textit{Embarrassment} depicted in the images. 

\subsection{Experiment C: Ablation Study on Environmental Contexts}

Experiment C studies how describing a person's environmental context in an image contributes to predicting a person's emotion. Environmental contexts can range from location and time to different types of animals and activities, and this information provides valuable insight into what a person may be feeling. Especially when facial and body features are missing in an image, we can rely on scene context to predict an emotion. Fig.~\ref{embarrassment} (c) and (d) show two examples where a person's face is not visible in the images, and therefore, scene context becomes important to accurately predict their emotion. 

This experiment used the same dataset as Experiment A, but all information about a person's environment and physical surrounding was removed from the caption. Therefore, the captions used in the experiment contained only perceived age, perceived sex, applicable social identity, face and body signals, and interactions and relationships with others. Table~\ref{caption_table} shows a caption without environmental context for Fig.~\ref{WebAnnotation}. 

% If we exclude the information that the girl in the red bounding box is standing at a trail intersection, then all we know about this image would be \textit{there is a girl}. It would be difficult for a large language model, and even for humans too, to infer what she is feeling just by looking at her. In the other image, if we ignore the dog that is jumping and biting the man's arm, then all the information that we have about him would be \textit{there is a man bending down}. Therefore, by adding contextual information about their physical surroundings such as "at a trail intersection" and "a dog jumping and biting", GPT-3 would be able to predict more accurately that the girl may be feeling \textit{Confusion}, not knowing which way to go, whereas the man may be feeling \textit{Fear} or \textit{Pain/Suffering - Physical} from the dog.

\begin{figure}[t]
\centerline{\includegraphics[scale=0.5]{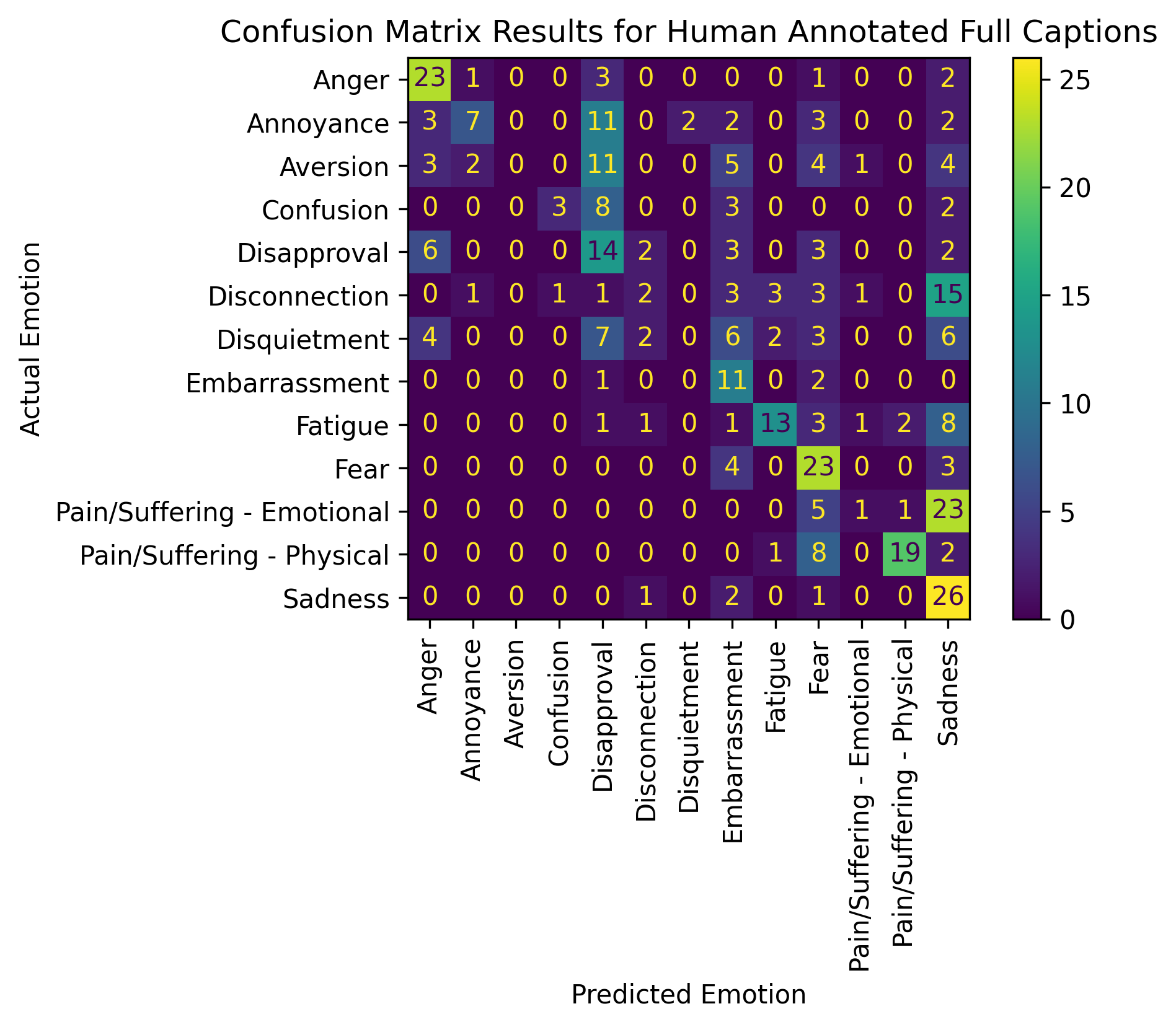}}
\caption{Confusion matrix from the experiment using full captions.}
\label{1_confusion_matrix}
\end{figure}

\begin{figure}[t]
\centerline{\includegraphics[scale=0.5]{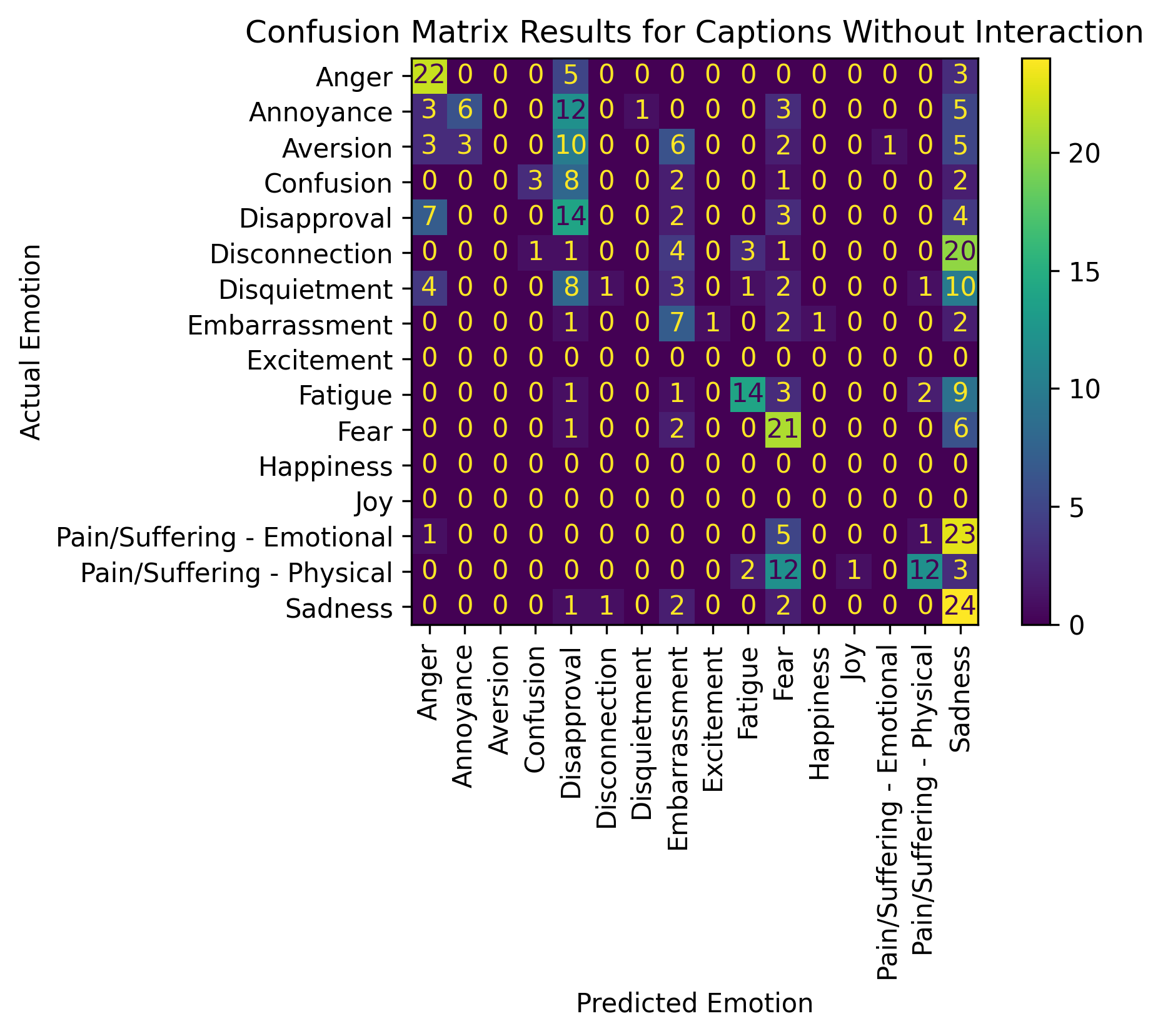}}
\caption{Three emotions - Excitement, Happiness and Joy, that were not on the list of emotions that we provided to GPT-3.5 to choose from, were predicted.}
\label{1b_confusion_matrix}
\end{figure}

\begin{figure}[t]
\centerline{\includegraphics[scale=0.5]{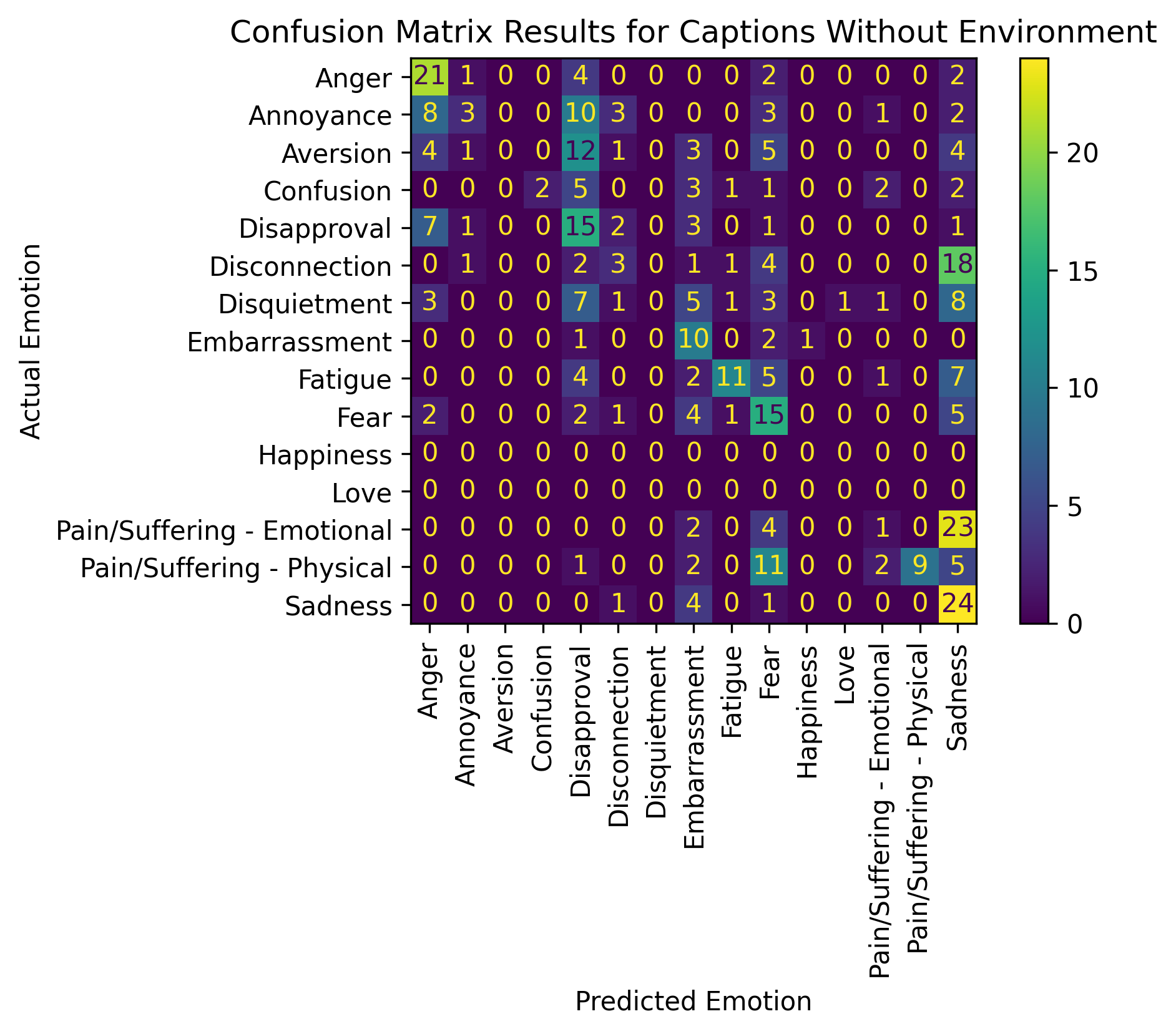}}
\caption{Two emotions - Happiness and Love, that were not on the list of emotions that we provided to GPT-3.5 to choose from, were predicted.}
\label{1c_confusion_matrix}
\vspace{-4mm}
\end{figure}

\begin{figure}[t]
\centerline{\includegraphics[width=9cm]{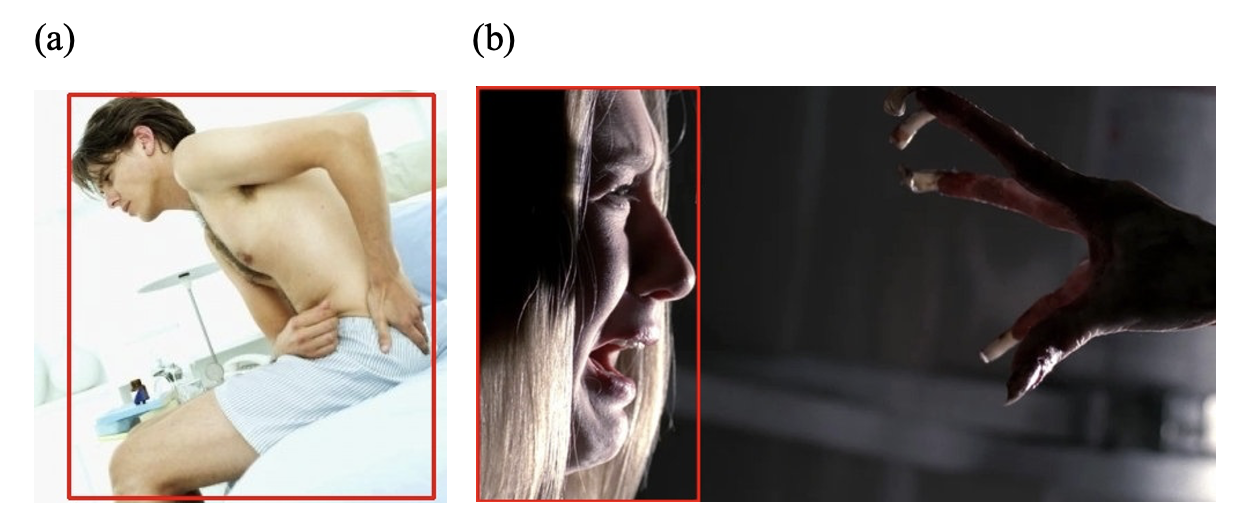}}
\caption{Image (a) shows that when \textit{medication on the side} was removed from the full caption, the predicted emotion changed from Physical Pain/Suffering to Disapproval. Image (b) shows that when \textit{right in front of an alien hand in the dark} was removed from the full caption, the predicted emotion changed from Fear to Disapproval.}
\label{Fear}
\vspace{-2mm}
\end{figure}

\begin{figure}[t]
\centerline{\includegraphics[width=8cm]{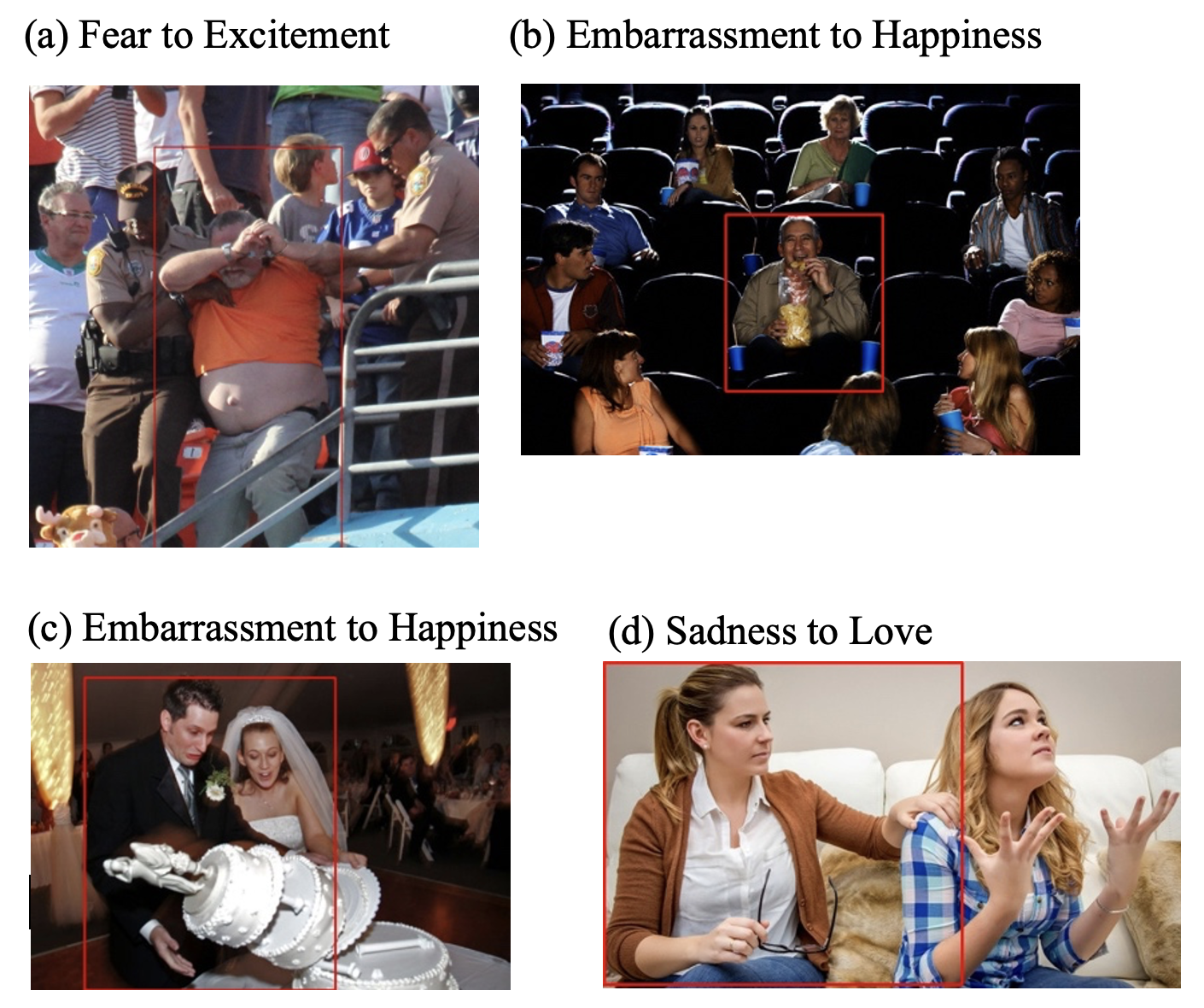}}
\caption{Examples of images where new positive emotions such as Excitement, Happiness, and Love were predicted by GPT-3.5 when interactions and environments were removed from the captions. The first emotion above each image was generated using full captions.}
\label{new_emotions}
\vspace{-2mm}
\end{figure}

\section{Results and Analysis}
The results of our GPT-3.5 emotion prediction are shown in Table ~\ref{stats_table}. The table contains precision, recall, and F1 score for each emotion and the total accuracy for each experiment. Experiment A with full captions has the highest accuracy. Experiment C with environments removed has the lowest accuracy. The confusion matrix results are in Fig.~\ref{1_confusion_matrix},
Fig.~\ref{1b_confusion_matrix}, and Fig.~\ref{1c_confusion_matrix}. From the confusion matrix for full captions in Fig.~\ref{1_confusion_matrix}, we can notice that \textit{Anger} and \textit{Sadness} were the most picked, but overall, according to F1 scores, \textit{Physical Pain/Suffering} was best estimated. GPT-3.5 was not able to predict \textit{Aversion}. \textit{Disconnection} and \textit{Disquietment} were also not well recognized. \textit{Emotional Pain/Suffering} was frequently recognized as \textit{Sadness}, which may be reasonable. 
\textit{Annoyance} and \textit{Confusion} were often recognized as \textit{Disapproval}. \textit{Fear} 
appeared to need environmental cues to be well predicted. \textit{Disapproval} and \textit{Fatigue} seem not to be impacted by social and environmental contexts. \textit{Embarrassment} was fairly well predicted with social interactions.

\subsection{Importance of Interactions in Emotion Estimation}
  
The lack of context about a person's social interactions with other people seems to impact how \textit{Embarrassment} was perceived by GPT-3.5 the most. Its F1 score for Experiment B without interactions is $0.33$, which is lower than the two experiments with interactions, $0.41$ and $0.38$. A potential reason is that for all the sample images that were annotated as \textit{Embarrassment} as shown in Fig.~\ref{embarrassment} (a) and (b), multiple people were present, and there were interactions between the people. In this case, removing social interaction contexts from the captions removes a critical piece of information that may indicate a person may be feeling embarrassed. 
% This is a limitation that we will discuss in the Limitations section. 

\textit{Physical Pain/Suffering} is another emotion that may need social interactions to be well recognized. Its F1 score dropped from $0.73$ with full captions to $0.52$ without interaction and was predicted as \textit{Fear} by GPT-3.5 more times when compared to full captions.

\subsection{Importance of Environments in Emotion Estimation}

The lack of description about environment in the image captions seemed to impact \textit{Physical Pain/Suffering} the most. The F1 score dropped from $0.73$ with full captions to $0.46$ and it is even lower than the F1 score for captions without interactions. As a result, we observed that \textit{Physical Pain/Suffering} needed both the interaction and environment descriptions to be well recognized by GPT-3.5, especially environments out of the two. An example is shown in Fig.~\ref{Fear} (a) where the predicted emotion changed from \textit{Physical Pain/Suffering} to \textit{Disapproval} after removing the environmental description. The full caption for this image with the removed part in italics is: Jack is a male adult. Jack is or has frowning, rubbing the back. \textit{Jack's physical environment is on a bed with medication on the side.}

Another emotion that may benefit from environmental context is \textit{Fear}. Its F1 score dropped from $0.52$ and $0.48$ with environments to $0.34$ without environments. An example is shown in Fig.~\ref{Fear} (b) where the predicted emotion changed from \textit{Fear} to \textit{Disapproval}. The full caption for this image with the removed part in italics is: Chloe is a female adult. Chloe is or has frowning, open mouth. \textit{Chloe's physical environment is right in front of an alien hand in the dark.}

\subsection{Importance of Facial and Body Signals}

Across the three experiments, we noticed \textit{Anger} was well recognized as \textit{Anger} by GPT-3.5 with F1 scores of $0.67$, $0.63$, and $0.56$. A reason for this may be that our samples had distinct facial expressions and body postures that differentiated it from the other emotions, like furrowed eyebrows, gritting teeth, and wrinkling nose.

GPT-3.5's understanding of \textit{Disapproval} also seems to not be affected by the existence of interaction and environment descriptions in the captions. The F1 scores are $0.32$, $0.31$, and $0.32$. This can suggest that \textit{Disapproval} was recognized through a person's facial expressions and body poses more than a person's physical surrounding. Body gestures such as crossed arms, pointing finger, and thumbs down may all be indicators of \textit{Disapproval} to GPT-3.5.

\textit{Sadness} was almost always predicted as \textit{Sadness}, but \textit{Emotional Pain/Suffering} was frequently predicted as \textit{Sadness} in all three experiments as well. In fact, the F1 scores for \textit{Emotional Pain/Suffering} are only $0.06$, $0$, and $0.05$. This may indicate that these two emotions share similar physical signals, such as crying, downturned mouth, and tilting head downward, and thus GPT-3.5 was not capable of differentiating the two emotions. Or, it could be that GPT-3.5 tends to select the more commonly known emotion from the list that it was provided.

\subsection{New Emotions Predicted by GPT-3.5}
Interestingly, four positive emotions were predicted but they were not on the list of 13 negative emotions that we provided to GPT-3.5 to choose from. The emotions and their number of occurrences are Excitement (1), Happiness (2), Joy (1), and Love (1). All four cases happened in the two ablation studies when interaction or environmental features were missing from the captions. Fig.~\ref{new_emotions} shows the image for these cases and the corresponding captions. We also show the emotions that were originally predicted using the full captions and the new emotions they were changed to. The removed interactions and environments are in italics:

\begin{itemize}
    \item Fig.~\ref{new_emotions} (a) \textbf{Fear to Excitement.} Terry is a male adult. \textit{Karl is a security guard and he is grabbing onto Terry and carrying him out from the stadium.} Terry's physical environment is at a sports game.
    \item Fig.~\ref{new_emotions} (b) \textbf{Embarrassment to Happiness.} Jack is a male adult. Jack is or has smiling. \textit{Beth is a customer and she is side-eyeing Jack. Zoe is a customer and she is staring at Jack.} Jack's physical environment is eating in a movie theatre.
    
    % \item (c) \textbf{Joy.} Sophia is a female adult. Sophia's physical environment is in a hospital bed and gave birth to a child.

    \item Fig.~\ref{new_emotions} (c) \textbf{Embarrassment to Happiness.} Lucas is a male adult. Lucas is a(n) groom. Lucas is or has lips that flatten, palms open. Mia is Lucas' bride and she is smiling. \textit{Lucas' physical environment is cake falling down at wedding.}
    \item Fig.~\ref{new_emotions} (d) \textbf{Sadness to Love.} Jane is a female adult. Jane is or has taking off eyeglasses. Mia is Jane's daughter and Jane is putting her hand on Mia's shoulder while Mia has her back turned to Jane. \textit{Jane's physical environment is on a couch.}
    
\end{itemize}

There was no new negative emotion.

\section{Discussion and Future Work}
We proposed a new approach for emotion estimation which couples text-based contextual descriptions of people in images with LLMs. Towards this goal, this study provided a benchmark of GPT-3.5 on a set of image captions depicting negative emotions. We also investigated the contributions of social cues and broader contextual information when perceiving human emotions. For example, we observed that \textit{Embarrassment} and \textit{Sadness} contained overlapping physical signals, such as ``covering own face" or ``tilting head downwards", and that social interactions could help distinguish between these two labels, i.e., a person covering their face with a tilted down head, along with being pointed at and laughed at by others, could appear to suggest Embarrassment. Moreover, our results showed that \textit{Aversion} was never predicted as an emotion across all three experiments on 360 captions, while \textit{Disquietment} was only predicted three times. \textit{Emotional Pain/Suffering} and \textit{Disconnection} were also frequently predicted as \textit{Sadness}, even in the presence of scene contexts. A possible explanation for this may be that GPT-3.5 might not have been sufficiently trained on language data that contained such emotion words.
 
  The study is not without limitations. Firstly, we only focused on the negative emotions of the EMOTIC dataset, and the social signals list employed for annotations was restricted to those associated with our set of negative emotions. Secondly, the list of social signals was partially generated by LLMs and subsequently tested on them.  The resulting contextual descriptions were ultimately determined and validated by our team of annotators. Aside from that, the size of the resulting descriptions list, as well as the number of annotated images, was relatively small. Finally, our study did not delve into the individual contribution of each physical description or demographic information for emotion detection, making it an interesting area to explore for future work. In the future, an independent perception study of the captions and ablations could also help provide a comparison to the GPT-3.5 results, as well as addressing the challenge of fully automatic captioning, and evaluating over all EMOTIC labels.

Overall, our approach may be used to enhance transparency and facilitate an effective breakdown of scene representation for contextual emotion estimation. It is hoped that our study can also serve as a catalyst for future research in interpretability of LLMs, as well as understanding human perception of emotions, especially if reproduced with other languages and cultures.

\section{Ethical Impact Statement}

\emph{Issues Related to Human Subjects.}
In this study, all social signal and context coding and emotion annotation of the images was performed by two members of the research team. The photos are from the EMOTIC dataset which contain images from the internet, of which some belong to the public datasets MSCOCO and Ade20k. Access to the EMOTIC dataset requires a request to the database authors. The research team are not related to the pictures. The images are of people who may be experiencing negative emotions including grief at a funeral, protesting, or war. It does not contain images  stronger than images that a person might encounter in news media (e.g. no nudity, torture, etc.)

\emph{Potential Negative Societal Impact.} Software that can detect negative emotions from images accurately could potentially be used for surveillance by authorities for intervention and restriction of autonomy. The application of this research for such use is not condoned by the authors.

\emph{Limits of Generalizability}.
The proposed list of physical signals is not claimed to be exhaustive, not only because we focus on a limited set of negative emotions, but also that different cultures will express emotions with different facial and bodily signals. The source Emotion Thesaurus is written by North American authors, and similar writing guides in other languages may produce differing results. ChatGPT, used to supplement the Emotion Thesaurus, also contains its own biases ~\cite{nadeem2020stereoset}. In addition, GPT-3.5, trained in English, carries biases in the association of facial, bodily and contextual signals with the final emotion. In addition, there were also only two annotators, and we acknowledge that they may also carry their own cultural bias.

\emph{Other Issues}. This work relied on a pre-trained large language model GPT-3.5. While this work did not perform any additional training, the carbon cost of training LLMs cannot be underestimated~\cite{patterson2021carbon}.

\renewcommand{\BibTeX}{\footnotesize}
{
%\small
%\bibliographystyle{IEEEtran}
\bibliographystyle{IEEEtran}
\bibliography{ref}

\begin{thebibliography}{10}
\providecommand{\url}[1]{#1}
\csname url@samestyle\endcsname
\providecommand{\newblock}{\relax}
\providecommand{\bibinfo}[2]{#2}
\providecommand{\BIBentrySTDinterwordspacing}{\spaceskip=0pt\relax}
\providecommand{\BIBentryALTinterwordstretchfactor}{4}
\providecommand{\BIBentryALTinterwordspacing}{\spaceskip=\fontdimen2\font plus
\BIBentryALTinterwordstretchfactor\fontdimen3\font minus
  \fontdimen4\font\relax}
\providecommand{\BIBforeignlanguage}[2]{{%
\expandafter\ifx\csname l@#1\endcsname\relax
\typeout{** WARNING: IEEEtran.bst: No hyphenation pattern has been}%
\typeout{** loaded for the language `#1'. Using the pattern for}%
\typeout{** the default language instead.}%
\else
\language=\csname l@#1\endcsname
\fi
#2}}
\providecommand{\BIBdecl}{\relax}
\BIBdecl

\bibitem{barrett2019emotional}
L.~F. Barrett, R.~Adolphs, S.~Marsella, A.~M. Martinez, and S.~D. Pollak,
  ``Emotional expressions reconsidered: Challenges to inferring emotion from
  human facial movements,'' \emph{Psychological Science in the Public
  Interest}, vol.~20, no.~1, pp. 1--68, 2019.

\bibitem{pantic2000expert}
M.~Pantic and L.~J. Rothkrantz, ``Expert system for automatic analysis of
  facial expressions,'' \emph{Image and Vision Computing}, vol.~18, no.~11, pp.
  881--905, 2000.

\bibitem{schindler2008recognizing}
K.~Schindler, L.~Van~Gool, and B.~De~Gelder, ``Recognizing emotions expressed
  by body pose: A biologically inspired neural model,'' \emph{Neural Networks},
  vol.~21, no.~9, pp. 1238--1246, 2008.

\bibitem{barrett2011context}
L.~F. Barrett, B.~Mesquita, and M.~Gendron, ``Context in emotion perception,''
  \emph{Current Directions in Psychological Science}, vol.~20, no.~5, pp.
  286--290, 2011.

\bibitem{barrett2017emotions}
L.~F. Barrett, \emph{How emotions are made: The secret life of the
  brain}.\hskip 1em plus 0.5em minus 0.4em\relax Pan Macmillan, 2017.

\bibitem{calbi2017context}
M.~Calbi, K.~Heimann, D.~Barratt, F.~Siri, M.~A. Umilt{\`a}, and V.~Gallese,
  ``How context influences our perception of emotional faces: A behavioral
  study on the kuleshov effect,'' \emph{Frontiers in Psychology}, vol.~8, p.
  1684, 2017.

\bibitem{7284842}
U.~Hess and S.~Hareli, ``The influence of context on emotion recognition in
  humans,'' in \emph{Proceedings of the 11th IEEE International Conference and
  Workshops on Automatic Face and Gesture Recognition (FG)}, vol.~03, 2015, pp.
  1--6.

\bibitem{kosti2019context}
R.~Kosti, J.~M. Alvarez, A.~Recasens, and A.~Lapedriza, ``Context based emotion
  recognition using {EMOTIC} dataset,'' \emph{IEEE Transactions on Pattern
  Analysis and Machine Intelligence}, vol.~42, no.~11, pp. 2755--2766, 2019.

\bibitem{le2022global}
N.~Le, K.~Nguyen, A.~Nguyen, and B.~Le, ``Global-local attention for emotion
  recognition,'' \emph{Neural Computing and Applications}, vol.~34, no.~24, pp.
  21\,625--21\,639, 2022.

\bibitem{mittal2020emoticon}
T.~Mittal, P.~Guhan, U.~Bhattacharya, R.~Chandra, A.~Bera, and D.~Manocha,
  ``Emoti{C}on: Context-aware multimodal emotion recognition using frege's
  principle,'' in \emph{Proceedings of the 2020 IEEE/CVF Conference on Computer
  Vision and Pattern Recognition}, 2020, pp. 14\,234--14\,243.

\bibitem{wang2022context}
Z.~Wang, L.~Lao, X.~Zhang, Y.~Li, T.~Zhang, and Z.~Cui, ``Context-dependent
  emotion recognition,'' \emph{Journal of Visual Communication and Image
  Representation}, vol.~89, p. 103679, 2022.

\bibitem{dudzik2020exploring}
B.~Dudzik, J.~Broekens, M.~Neerincx, and H.~Hung, ``Exploring personal memories
  and video content as context for facial behavior in predictions of
  video-induced emotions,'' in \emph{Proceedings of the 2020 International
  Conference on Multimodal Interaction}, 2020, pp. 153--162.

\bibitem{shin2022contextual}
S.~Shin, D.~Kim, and C.~Wallraven, ``Contextual modulation of affect: Comparing
  humans and deep neural networks,'' in \emph{Companion Publication of the 2022
  International Conference on Multimodal Interaction}, 2022, pp. 127--133.

\bibitem{vaswani2017attention}
A.~Vaswani, N.~Shazeer, N.~Parmar, J.~Uszkoreit, L.~Jones, A.~N. Gomez,
  {\L}.~Kaiser, and I.~Polosukhin, ``Attention is all you need,''
  \emph{Advances in Neural Information Processing Systems}, vol.~30, 2017.

\bibitem{radford2019language}
A.~Radford, J.~Wu, R.~Child, D.~Luan, D.~Amodei, and I.~Sutskever, ``Language
  models are unsupervised multitask learners,'' \emph{OpenAI blog}, vol.~1,
  no.~8, p.~9, 2019.

\bibitem{devlin2018bert}
J.~Devlin, M.-W. Chang, K.~Lee, and K.~Toutanova, ``Bert: Pre-training of deep
  bidirectional transformers for language understanding,'' \emph{arXiv preprint
  arXiv:1810.04805}, 2018.

\bibitem{brown2020language}
T.~Brown, B.~Mann, N.~Ryder, M.~Subbiah, J.~D. Kaplan, P.~Dhariwal,
  A.~Neelakantan, P.~Shyam, G.~Sastry, A.~Askell \emph{et~al.}, ``Language
  models are few-shot learners,'' \emph{Advances in Neural Information
  Processing Systems}, vol.~33, pp. 1877--1901, 2020.

\bibitem{antol2015vqa}
S.~Antol, A.~Agrawal, J.~Lu, M.~Mitchell, D.~Batra, C.~L. Zitnick, and
  D.~Parikh, ``{VQA}: Visual question answering,'' in \emph{Proceedings of the
  2015 IEEE International Conference on Computer Vision}, 2015, pp. 2425--2433.

\bibitem{vinyals2015show}
O.~Vinyals, A.~Toshev, S.~Bengio, and D.~Erhan, ``Show and tell: A neural image
  caption generator,'' in \emph{Proceedings of the 2015 IEEE Conference on
  Computer Vision and Pattern Recognition}, 2015, pp. 3156--3164.

\bibitem{puglisi2019emotion}
B.~Puglisi and A.~Ackerman, \emph{The emotion thesaurus: A writer's guide to
  character expression}.\hskip 1em plus 0.5em minus 0.4em\relax JADD
  Publishing, 2019, vol.~1.

\bibitem{nadeem2020stereoset}
M.~Nadeem, A.~Bethke, and S.~Reddy, ``{S}tereo{S}et: Measuring stereotypical
  bias in pretrained language models,'' \emph{arXiv preprint arXiv:2004.09456},
  2020.

\bibitem{patterson2021carbon}
D.~Patterson, J.~Gonzalez, Q.~Le, C.~Liang, L.-M. Munguia, D.~Rothchild, D.~So,
  M.~Texier, and J.~Dean, ``Carbon emissions and large neural network
  training,'' \emph{arXiv preprint arXiv:2104.10350}, 2021.

\end{thebibliography}
}

\end{document}